\documentclass[default,iicol]{sn-jnl}

\jyear{2021}%

\theoremstyle{thmstyleone}%
\theoremstyle{thmstyletwo}%

\theoremstyle{thmstylethree}%
\usepackage{xcolor}

\usepackage{comment}

\begin{document}

\title[Article Title]{Regularization-based Pruning of Irrelevant Weights in Deep Neural Architectures}


\author*[1]{\fnm{Giovanni} \sur{Bonetta}}\email{giovanni.bonetta@unito.it}

\author[1]{\fnm{Matteo} \sur{Ribero}}\email{matteo.ribero@edu.unito.it}

\author[1]{\fnm{Rossella} \sur{Cancelliere}}\email{rossella.cancelliere@unito.it}

\affil[1]{\orgdiv{Computer Science Department}, \orgname{University of Turin}, \orgaddress{\street{Via Pessinetto}, \city{Turin}, \postcode{10149}, \state{Italy}}}



\abstract{Deep neural networks exploiting million parameters are currently the norm. This is a potential issue because of the great number of computations needed for training, and the possible loss of generalization performance of overparameterized networks. We propose in this paper a method for learning sparse neural topologies via a regularization approach that identifies nonrelevant weights in any type of layer (i.e., convolutional, fully connected, attention and embedding ones) and selectively shrinks their norm while performing a standard back-propagation update for relevant layers. 
This technique, which is an improvement of classical weight decay, is based on the definition of a regularization term that can be added to any loss function regardless of its form, resulting in a unified general framework exploitable in many different contexts. The actual elimination of parameters identified as irrelevant is handled by an iterative pruning algorithm. 

To explore the possibility of an interdisciplinary use of our proposed technique, we test it on six different image classification and natural language generation tasks, among which four are based on real datasets. We reach state-of-the-art performance in one out of four imaging tasks while obtaining results better than competitors for the others and one out of two of the considered language generation tasks, both in terms of compression and metrics.
}




\maketitle

\section{Introduction}
\label{sec:intro}
Deep learning models have consistently established in the past few years new state-of-the-art performances in a flood of different domains, including image processing \cite{Zhang20,He19,He2016,Szegedy15}, image captioning \cite{Guo20,Feng19}, language generation \cite{Puduppully19,Dusek20}, and machine translation \cite{Vaswani2017,Bahdanau14}.

The resources required to properly train them, however, can be prohibitive, since the number of weights used for these tasks may easily sum up to several million. These growing performance costs have therefore induced scientists to look for techniques limiting the size of neural architecture parameters. 

An effective approach for reducing this complexity is sparsity, defined as the property that a substantial subset of the model weights have a value of zero (i.e., layers’ weights matrices are sparse). Sparsity allows smaller computational and storage requirements, and as shown, for example, in \cite{Han2015} and \cite{Ullrich17}, deep architectures tolerate it well.

It can shorten training time and reduce the memory footprint of regular networks to fit mobile devices, at only a small cost in accuracy. Smaller models are easier to send on edge devices and are also significantly less energy greedy, as noted in  \cite{Sanh}: ``the majority of the energy consumption comes from fetching the model parameters from the long term storage of the mobile device to its volatile memory".
In addition, sparsity is also a solution for improving inference performance through overfitting control and, as suggested by  \cite{Liu2017}, may lead to improved performance in transfer learning scenarios. 

Two main sparsity-inducing approaches can be found in the literature, referred to as unstructured or structured, depending on whether single weights or entire structured groups of weights and/or neurons are removed.

Numerous methods have been proposed over the past few years to reach these goals: a nonexhaustive review of the recent relevant literature can be found in Section ~\ref{sec:related}. 

The $ L2 $ regularization-based techniques, detailed in Section~\ref{sec:theory}, are among the most popular: they add a penalty term to the cost function to shrink the parameter values. All parameters dropping below a predefined threshold are then set to zero, thus obtaining sparse architectures. 

A drawback of these methods is that neural weights’ norms are all driven close to zero without accounting for weight relevance in the neural architecture, as discussed in detail in  \cite{tartaglione1,tartaglione2}.

Our work relies on this approach: we propose a new loss function holding a suited regularization term that is a specialization of the well-known weight decay. It allows the derivation of a new weight update rule that selectively decreases the norm of nonrelevant weights, while performing a standard backpropagation update for relevant weights. Weights’ update directly follows from loss optimization, not requiring the definition of ad hoc update rules, as frequently done (see \cite{Han2015,tartaglione1,tartaglione2,Gomez19}).
Decreased weights are then pruned to sparsify the neural architecture.

Our technique is general, as the proposed regularization term can be added to any loss function regardless of its form and constitutes a unified framework potentially exploitable for many different applications.

We verify the effectiveness of our method in the context of image classification and natural language generation, sparsifying convolutional (LeNet-5, ResNet32/50) and self-attention transformer-based neural architectures, respectively. While the former task has already been addressed in the literature, it is still rare to find sparsity techniques applied to language generation architectures.

We reach state-of-the-art results in one out of four image classification tasks, while establishing, to the best of our knowledge, new state-of-the-art performance for the others and one out of two of the considered language generation tasks. 

The rest of this paper is organized as follows: Section~\ref{sec:related} contains an overview of related works concerning structured and unstructured pruning. In Section~\ref{sec:theory}, the theoretical foundations of our model are presented, and Section~\ref{general_approach_pruning} outlines the details of our pruning algorithm. Sections~\ref{sec:imaging} and \ref{sec:language} describe the datasets, implementation details and results obtained in both contexts.

\section{Related Works}
\label{sec:related}

In this section, we outline some recent approaches investigating structured and unstructured sparsity techniques. Although our paper proposes an unstructured pruning method, for the sake of completeness and for introducing some works used for comparison in Section~\ref{sec:imaging}, we briefly resume in the following a few recent methods for structured sparsity.

In this context, block/layer pruning, group/filter/channel pruning, and kernel pruning approaches play an important role, particularly for convolutional-based deep networks.

Block/layer pruning (\cite{Wang19,Ding20}) aims at shrinking a network by removing entire blocks or layers; in particular, group pruning techniques are used to eliminate redundant groups. A finer-grained method is filter/channel pruning (\cite{Huang18,Lin18,Lin20}),which is used to eliminate redundant filters. Thus, the dimensionality of the feature maps is reduced, to the extent that even entire channels can be discarded.

Kernel pruning (\cite{Zhong18,Zhu22a,Zhu22b})aims at removing the basic feature extraction units, the k x k matrices (kernels) within filters, generating in this way sparse networks with a fine granularity. Lin et al. \cite{Zhong18} added a regularization term during training for pruning the connections characterized by less synaptic strength. Zhu and Pei  \cite{Zhu22a} proposed progressive pruning with saliency mapping of input‒output channels for solving the problem of missing channels during pruning and improving efficiency.

In the context of unstructured pruning, methods based on merely removing small norm weights are the simplest \cite{Han2015}. 

Methods grounded in Bayesian statistics constitute another possibility for achieving sparsity: the soft weight sharing (SWS) approach \cite{Ullrich17} succeeds in reducing the number of parameters to be stored, sharing the redundant parameters among layers. Weights are represented as a mixture of Gaussian distributions, where the components of the mixture are learned.

Sparse variational dropout \cite{Molchanov17} treats dropout as noise applied to the inputs of all neural layers; it represents the first attempt to use the dropout technique for sparsifying networks, paving the way for new research in the field (see, for instance,  \cite{alehinejad2021}). Targeted dropout \cite{Gomez19} is another dropout-based approach that focuses on disentangling an important subset of weights from unimportant weights to obtain stable pruning. Before computing the gradients for each weight update, targeted dropout stochastically selects a set of units or weights to be dropped, using the L1 and L2 norms as a proxy for weight importance, and then computes the gradients for the remaining weights.

The authors of SNIP \cite{Lee2019} define a saliency-based criterion on network connections to select the most important connections for the given task. Then, they prune the network accordingly in a one-shot way and finally train the model to obtain good accuracy. The drawback of this algorithm is that its one-shot nature prevents it from achieving a sparsity level comparable to iterative approaches.

Guo et al. attempted to address this issue by proposing the DNS  \cite{Guo16} technique, where incorrectly pruned connections can be reestablished by a splicing pass if evaluated as important, making the pruning effort dynamic.

As stated in Section~\ref{sec:intro}, regularization can be used as a particularly effective shrink-and-prune approach to sparsity \cite{Gale2019}but with the drawback that all weights are indiscriminately penalized.

An effective method for avoiding this issue is presented in  \cite{tartaglione1}, where the idea of output-based sensitivity is introduced: weights are selectively penalized depending on their capability to induce variations in network outputs when changed. A refinement of this method is presented in \cite{tartaglione2}, where state-of-the-art results are reached in image classification thanks to the introduction of loss-based sensitivity, which aims at shrinking weights that contribute the least to the final loss value.

One issue often noted with unstructured techniques is that the obtained sparse network has an irregular distribution of null weights, requiring specific software/hardware support for acceleration.
For this reason, frequently no significant improvements in inference speed are observed currently, but the topic remains interesting since hardware manufacturers are developing new chips designed to exploit the unstructured pattern\footnote{https://www.graphcore.ai/}.

\section{The Relevance-based Pruning Method: Theory}
\label{sec:theory}







Regularization methods turn an original unstable, ill-posed problem into a well-posed problem; they limit the capacity of models by adding a parameter norm penalty to the loss functional L, to control overfitting and decrease the test error (see \cite{Goodfellow2016}).

One of the most common parameter norm penalties is L2, commonly known as Tikhonov regularization \cite{tikhonov1}, ridge regression or weight decay. 

In a neural context, the regularized loss $\tilde{L} $ depends on each weight $ w^n_{i, j} $ belonging to layer $n$ and connecting neurons $ i$ and $ j$ and has the form:

\begin{equation}
\tilde{L}(\bar{w})=L(\bar{w})+\lambda\|\bar{w}\|^{2}=L(\bar{w})+\lambda\sum_{n, i, j}\lvert w^n_{i, j} \rvert^{2}
\label{eq:loss2}
\end{equation}

\noindent where $\bar{w}$ is the vector whose elements are $ w^n_{i, j} $ and $\lambda$ is the regularization parameter.
The iterative application of the stochastic gradient descent algorithm (SGD) at time step  $t$ 

\begin{equation}\label{eq:SGD}
 w^n_{i, j}(t) \equiv w^n_{i, j}(t-1) -\eta
    \frac{\partial \tilde{L}(\bar{w})}{\partial w^n_{i, j}} 
\end{equation}
results in the well-known weight decay update rule

\begin{equation}
\label{eq:loss3}
w^n_{i, j}(t)=w^n_{i, j}(t-1)-\eta \frac{\partial L(\bar{w})}{\partial w^n_{i, j}} -2\lambda w^n_{i, j}
\end{equation}

where $\eta$ is the learning rate.
The neural weights, therefore, are driven close to zero without taking account of their relevance in the neural architecture.

The main contribution of this work is the proposal of a new loss functional $ \hat{L} $, modified with respect to eq.~(\ref{eq:loss2}), allowing us to obtain a new selective weight decay rule that shrinks the magnitude of only those weights that are not relevant to the final error. 
We propose modifying the regularization term in eq.~(\ref{eq:loss2}) by multiplying it by a coefficient that measures how much the final loss value is influenced by modifying $  w^n_{i, j} $.

The quantity $\lvert\frac{\partial {L}}{\partial w^n_{i, j}}\rvert$ would seem to be a good candidate for this: small derivative values, for example, indicate that even a large variation in $ w^n_{i, j}$ does not cause large loss variations, i.e., weight changes are not relevant to the final loss value. In addition, large derivative values are not interesting because they characterize relevant weights, and we aim to drive to zero (and prune) only irrelevant weights. This derivative is not upper bounded, which is a possible issue in preserving convergence properties. 

Considering these requests, we therefore define the \emph{coefficient of irrelevance} $ {I}_{n, i, j} $ as:

\begin{equation}
{I}_{n, i, j} \equiv \exp (-\lvert\frac{\partial L}{\partial w^n_{i, j}}\rvert),\:\:\:\:\: 0<{I}_{n, i, j} <1 .
\end{equation}

 $ {I}_{n, i, j} $ is bounded and assumes values near 1 for irrelevant weights and near 0 for relevant weights. Moreover, it has the useful property to be almost everywhere differentiable.

We can now define the new loss functional $ \hat{L}$, modified with respect to eq.~(\ref{eq:loss2}) to selectively limit the magnitude of weights:
\begin{align}
 \hat{L}(\bar{w}) & \equiv{L}(\bar{w})+\lambda \sum_{n, i, j}({I}_{n, i, j}\cdot\lvert w^n_{i, j} \rvert^{2}) = \nonumber\\ 
 &= {L}(\bar{w}) +\lambda \sum_{n, i, j}(\exp(-\lvert \frac{\partial{L}}{\partial w^n_{i, j}}\rvert)\cdot \lvert w^n_{i,j} \rvert^{2}) \label{eq:mainloss}
\end{align}     

 The iterative application of stochastic gradient descent algorithm to  $ \hat{L}$ allows us to derive the new weights' update rule:

\begin{align}
& w^n_{i, j}(t) \equiv w^n_{i, j}(t-1) -\eta\frac{\partial \hat{\operatorname{L}}}{\partial w^n_{i, j}} = \nonumber \\&= w^n_{i, j}(t-1) -\eta \frac{\partial {\operatorname{L}}}{\partial w^n_{i, j}} -2\eta\lambda \exp (-\lvert\frac{\partial {L}}{\partial w^n_{i, j}}\rvert) w^n_{i, j} \nonumber\\& -\eta\lambda \lvert w^n_{i, j}\rvert^{2} \cdot \exp (-\lvert\frac{\partial {L}}{\partial w^n_{i, j}}\rvert) (-1) \frac{\partial}{\partial w^n_{i, j}}\lvert\frac{\partial {L}}{\partial w^n_{i, j}}\rvert = \nonumber \\ &= w^n_{i, j}(t-1) -\eta \frac{\partial {\operatorname{L}}}{\partial w^n_{i, j}} -2 \eta\lambda \exp (-\lvert\frac{\partial {L}}{\partial w^n_{i, j}}\rvert) w^n_{i, j} \nonumber\\ &+\eta\lambda \lvert w^n_{i, j}\rvert ^{2} \cdot \exp (-\lvert\frac{\partial {L}}{\partial w^n_{i, j}}\rvert) \text{sgn} (\lvert\frac{\partial^2{L}}{\partial {w^n_{i, j}}^2}\rvert ) \label{eq:demo1}
\end{align}

As usually done in first-order derivative optimization methods, we can neglect the second-order derivative term, so that eq.~(\ref{eq:demo1}) becomes:

\begin{equation}
 w^n_{i, j}(t) = w^n_{i, j}(t-1) -\eta
\frac{\partial {\operatorname{L}}}{\partial w^n_{i, j}} -2\eta\lambda \exp (-\lvert\frac{\partial {L}}{\partial w^n_{i, j}}\rvert) w^n_{i, j} \label{eq:demo2}
\end{equation}

Different weight updates are made in the two cases determined by the extreme values of ${I}$:
        \begin{itemize}
             \item $I \sim 0$: in this case, the weight is relevant. The third term in eq.~(\ref{eq:demo2}) is approximately zero, so a standard update is performed, without a targeted reduction in the weight norm..
            \item $I \sim 1$: in this case, the weight is irrelevant. Because ${I} \sim 1$ implies $ \frac{\partial {L}}{\partial w^n_{i, j}} \sim 0 $, the second term in eq.~(\ref{eq:demo2}) is approximately zero, and the update rule becomes:
            \begin{equation}
               w^n_{i, j}(t) \simeq w^n_{i, j}(t-1) -2\eta\lambda w^n_{i, j}
            \end{equation}
        \end{itemize}
We can see that in this case, the weight is actually driven to zero at each iteration, because if  $w^n_{i, j} >0$ then $\Delta w^n_{i, j} <0$ (i.e., the norm of a positive irrelevant weight is decreased); otherwise, if  $w^n_{i, j} <0$ then $\Delta w^n_{i, j} >0$.

For a better understanding of how the coefficient of irrelevance I behaves and affects the weight’s norm, we report here as an example the histograms of initial and final (i.e., after training) distributions of these two quantities for two layers of the convolutional architecture ResNet50, which is analyzed in detail in Section~\ref{ResNet50 on ImageNet}.

\begin{figure*}[]
\centering
\begin{minipage}{.5\textwidth}
  \centering
  \includegraphics[width=1.\linewidth]{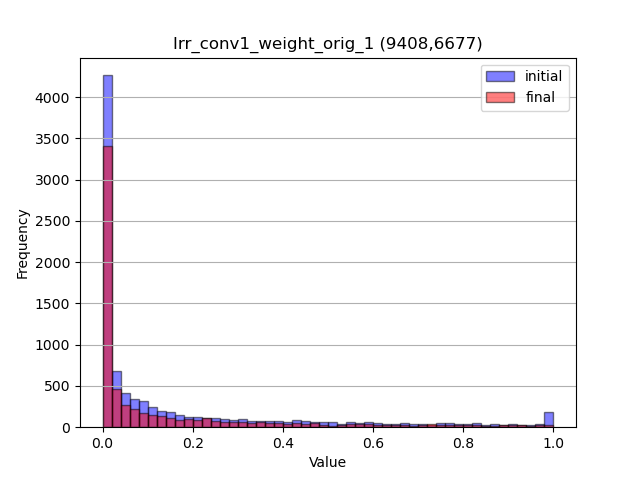}
  \caption{Irrelevance coefficient distribution (Conv1 layer).}
  \label{fig:Irr_conv1}
\end{minipage}%
\begin{minipage}{.5\textwidth}
  \centering
  \includegraphics[width=1.\linewidth]{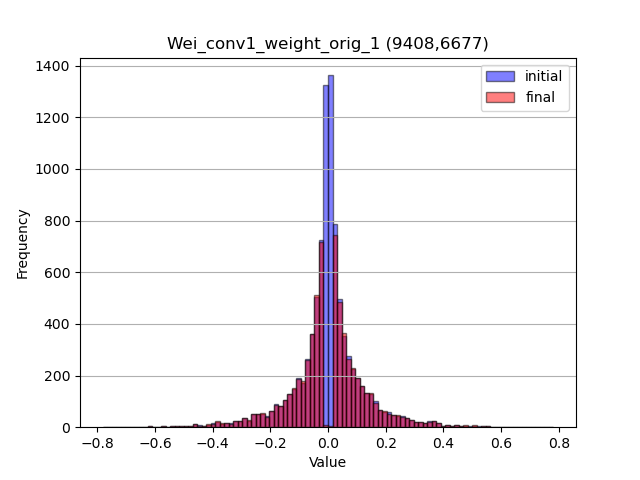}
  \caption{Weight norm distribution (Conv1 layer). There are 6,677 remaining weights, out of 9,408.}
  \label{fig:Wei_conv1}
\end{minipage}
\end{figure*}

Figure~\ref{fig:Irr_conv1} shows the histogram of the irrelevance coefficient values of the weights belonging to the first convolutional layer (which contains 9,408 weights) at the beginning (blue) and end (red) of the training of this model on the ImageNet dataset.  In this case, the vast majority of the weights have very low $ I $ I values for the entire training process, meaning that the corresponding weights are actually relevant and should not be pruned. Figure~\ref{fig:Wei_conv1} shows the distribution of the remaining weights’ norm: note that just 2,731 elements are pruned out of 9,408, which corresponds to 29\% of the layer weights. This number is far below 73.85\%, the percentage of pruned weights on the entire model  (as reported in Table \ref{tab:resnet50-imagenet-results}), and shows how relevant weights are less likely to be pruned than the other weights.

\begin{figure*}
\centering
\begin{minipage}{.5\textwidth}
  \centering
  \includegraphics[width=1.\linewidth]{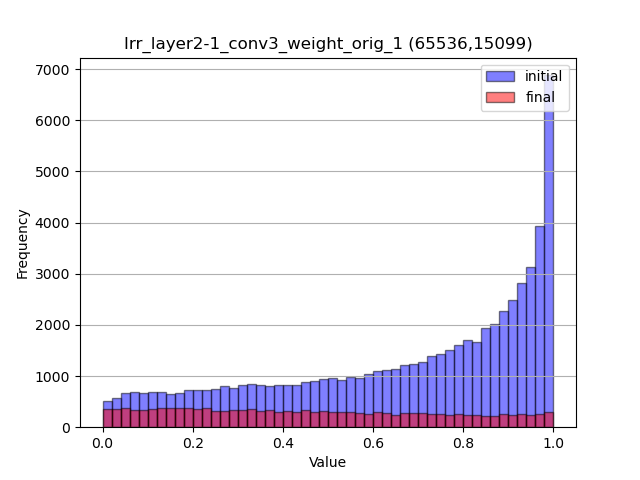}
  \caption{Irrelevance coefficient distribution (Conv18 layer).}
  \label{fig:Irr_layer2}
\end{minipage}%
\begin{minipage}{.5\textwidth}
  \centering
  \includegraphics[width=1.\linewidth]{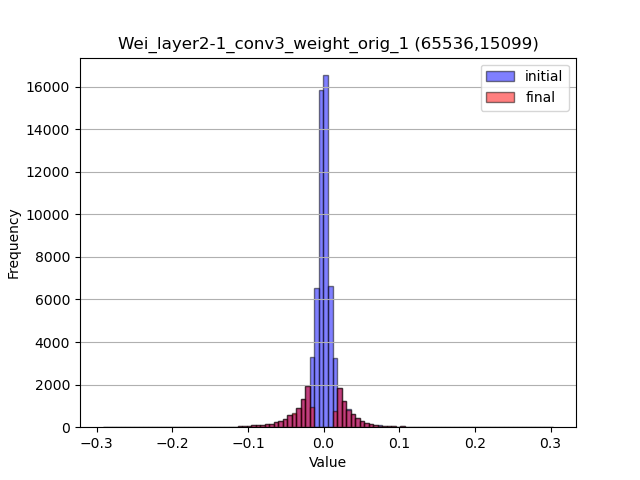}
  \caption{Weight norm distribution (Conv18 layer). There are 15099 remaining weights, out of 65536.}
  \label{fig:Wei_layer2}
\end{minipage}
\end{figure*}

Figures~\ref{fig:Irr_layer2} shows the histograms of the irrelevance coefficient values for the $18^{th}$ convolutional layer in the model. In this case, the majority of the I values at the beginning of the training is near 1, meaning that the corresponding weights are irrelevant and can be pruned.  As seen in Figure~\ref{fig:Wei_layer2}, the height of the final (red) weight histogram is dramatically reduced with respect to the initial (blue) histogram because training caused a dramatic reduction in the number of weights, up to 23.0\% of the initial weights. In addition, observing the final distribution of irrelevance coefficients values, we can note that it is definitely more uniform, meaning that the number of irrelevant weights is greatly decreased in percentage.

\section{Pruning Algorithm Description}
\label{general_approach_pruning}
        The pruning algorithm proceeds as follows:
        \begin{enumerate}
        \item We obtain a checkpoint from which to fine-tune. A common choice is to either find it in the literature or train it on our own. Another possibility is to use a randomly initialized checkpoint. 
        \item	We fine-tune the checkpoint (or train it from scratch when starting from a random initialization) by using our proposed regularization term, as in Equation  (\ref{eq:mainloss}). In this step, any optimizer can be used. During fine-tuning, we evaluate the model performance on the validation set each \textit{evaluation-interval} steps and:
        \begin{itemize}
        \item \textbf{prune}. If the validation performance is higher than a user-defined \textit{lower-bound}, a fixed percentage (\textit{pruning-percentage}) of the remaining model parameters is pruned, choosing from the ones with lower magnitude.
        \item \textbf{not prune}. If the validation performance is lower than the user-defined \textit{lower-bound}, the model is not ready to be pruned, so the fine-tuning proceeds normally.
        \end{itemize}
        \item These last two steps of the fine-tuning process are iteratively repeated until the model reaches a validation performance plateau (so it cannot be pruned further).
        \item We perform a final fine-tuning phase without regularization, aiming to obtain the best performing checkpoint.
        \end{enumerate}
When advanced in training, the model usually reaches an accuracy plateau, making it difficult for the algorithm to hit a pruning step. Because of this, we introduce an exponential decay schedule on  $\lambda$ that decreases the regularization term between the two validation steps. This procedure favors accuracy with respect to sparsification and helps the model to break the performance plateau and to cross over the lower bound, leading to further pruning.

Some hyperparameters are needed to implement this process:
\begin{itemize}
\item \textit{evaluation interval}: number of steps between two validation performance assessments.
\item \textit{lower bound}: performance lower bound. It is chosen as slightly lower than the state-of-the-art performance. For the image classification tasks, the performance is measured by accuracy, while for the language generation tasks, we use the metric BLEU \cite{Papineni2002}, which is briefly introduced in  Section~\ref{sec:language}.
\item \textit{pruning percentage}: The percentage of remaining nonzero parameters to be pruned at every pruning step.
\end{itemize}
If not stated otherwise, hyperparameters are chosen via grid search both in the training and fine-tuning phases.

We performed all our experiments using four Nvidia TITAN RTX 24Gb GPU, and the code is available at \url{https://github.com/giobin/Applied_Intelligence_sparsity}.

\section{Imaging}
\label{sec:imaging}

We test our method on four different image classification datasets chosen among the most popular benchmarks in the literature for sparsity research. Each of them is processed using architectures producing state-of-the-art performance.

\subsection{Dataset Description}

    \paragraph{MNIST}
        \label{MNIST}
        MNIST \cite{lecun2010}, is composed of 70,000 28x28 grayscale images containing handwritten numbers; the dataset is split into a training set (60,000 images) and a test set (10,000 images).

    \paragraph{Fashion-MNIST}
        Fashion-MNIST \cite{HanXiao2017}, based on the assortment of Zalando’s website, is a dataset comprising 28×28 grayscale images of 70,000 fashion products from 10 categories, with 7,000 images per category. The training set has 60,000 samples, and the test set has 10,000, and, although similar to MNIST, is more challenging.

    \paragraph{CIFAR-10}
      CIFAR-10 \cite{CIFAR10}, from the Canadian Institute for Advanced Research, is a subset of the Tiny Images dataset \cite{Torralba2008} and consists of 60,000 32x32 color images labeled with one of 10 mutually exclusive classes: airplane, automobile, bird, cat, deer, dog, frog, horse, ship, and truck. There are 6,000 samples per class, split into 5,000 for training and 1,000 for testing.

    \paragraph{ImageNet} 
         ImageNet \cite{deng2009imagenet} is arranged according to the WordNet \cite{Miller1995} noun hierarchy and is a real image database, in which each node in the hierarchy is represented by thousands of images; it contains more than 20,000 categories. In total, 14 million pictures were hand-annotated, and for one million of those, the bounding boxes were also provided. The RGB images have an average size of 469x387 pixels but are usually preprocessed by sampling them at 256x256 pixels.

\begin{table}[]
\caption{Hyperparameters used in imaging experiments.}
\label{tab:image hyperparameters}
\resizebox{\linewidth}{!}{%
\begin{tabular}{llll}
Hyperparameters    & MNIST/ & CIFAR-10 & ImageNet \\ 
    & Fashion-MNIST &  &  \\ \hline
\# epochs          & 120 & 290 & 40                     \\
\# batch size      & 100 & 128 & 200                       \\
$\eta$                & 0.001 & 0.0005 & 0.0001                        \\
Optimizer       & Adam & SGD & SGD\\
lower-bound       & 98.7 & 92.9 & 75.0                          \\
$\lambda$            & 0.001 & 1e-6 & 1e-4                        \\
pruning-percentage & 4\% & 4\% & 20\%                          \\
eval-interval      & 250 & 25 & 500                         \\ \hline
\end{tabular}%
}
\end{table}

\subsection{LeNet-5 on MNIST}
\label{LeNet-5 on MNIST}
Detail regarding LeNet-5 architecture are given in Table \ref{tab:image hyperparameters}.
This network \cite{LeCun89} consists of a convolution (Conv) layer with 6 5x5 filters, a 2x2 pooling layer, a convolution layer with 10 5x5 filters, another 2x2 pooling layer and three fully connected (FC) layers (120, 84, 10), for a total of 431,080 parameters. 

Since MNIST is an easy dataset, a pretrained checkpoint is not necessary, so we directly trained with regularization from scratch, using hyperparameter values resumed in Table \ref{tab:image hyperparameters}. Finally, we fine-tuned without regularization for 5 additional epochs.

The results from our model and competitors are shown in Table \ref{tab:lent5_MNIST_results}, together with the performance of the nonsparsified baseline model. The ``Sparsity(\%)" column refers to the percentage of pruned weights of each model with respect to the total number of baseline weights. The ``Compression Ratio" column is the ratio between the total number of weights in the baseline model and the number of remaining weights after pruning.

\begin{table*}[]
\centering
\caption{Test results for the LeNet-5 architecture on the MNIST dataset. (var) is the variance computed over 10 runs.}
\label{tab:lent5_MNIST_results}
\resizebox{\textwidth}{!}{
\begin{tabular}{lllllccc}
\hline
                                                        & \multicolumn{4}{c}{Residual Weights (\%)} & $\text{Accuracy}_{(var)}$ (\%) & Sparsity (\%) & \multicolumn{1}{c}{\begin{tabular}[c]{@{}c@{}}Compression\\ Ratio\end{tabular}} \\
Methods                                                 & Conv1     & Conv2     & FC1     & FC2     &               &               &                                                                                \\
Baseline                                                & 100       & 100       & 100     & 100     & 99.32         & --            &                                                                                \\
                                                        &           &           &         &         &               &               &                                                                                \\
Han et al., 2015 \cite{Han2015}        & 66        & 12        & 8       & 19      & 99.23         & 91.59         & 11.9x                                                                          \\
Tart. et al., 2018 \cite{tartaglione1} & 67.6      & 11.8      & 0.9     & 31.0    & 99.22         & 98.04         & 51.0x                                                                          \\
DNS \cite{Guo16}                       & 14        & 3         & 0.7     & 4       & 99.09         & 99.09         & 109.8x                                                                         \\
SWS \cite{Ullrich17}                   & -         & -         & -       & -       & 99.03         & 99.38         & 161.3x                                                                         \\
Tart. et al., 2021 \cite{tartaglione2} & 22        & 2.38      & 0.22    & 5.98    & 99.21         & 99.56         & 222.2x                                                                         \\
L0 \cite{Louizos18}                    & 45        & 36        & 0.4     & 5       & 99.00         & 98.57         & 69.9x                                                                          \\
Sparse VD \cite{Molchanov17}           & 33        & 2         & 0.2     & 5       & \textbf{99.25}        & 99.64         & 277.7x                                                                         \\
Our method                                              & 29        & 1.82      & 0.11    & 3.35    & $99.23_{\:(0.001)}$         & \textbf{99.71}         & \textbf{344.8x} \\ \hline
\end{tabular}
}
\end{table*}

Performances on MNIST are very similar to each other since accuracy and sparsification on this simple task reached the top possible, i.e., very close to 100\%. With our method, we obtain fewer than 1.5 k nonzero residual weights, a result primarily due to a better sparsification, when compared to the other works, of the fully connected layers, where the majority of the weights are. We obtain almost the best accuracy with the only exception of Sparse VD, even though it has higher sparsity. We also note that we obtain the same result as Han et al. but with  ~8\% fewer weights. 

Table \ref{tab:models_compression} shows the disk space occupied by the pruned and unpruned models after being compressed using the GZIP\footnote{more info at:\url{https://www.gnu.org/software/gzip/}} and BZIP2\footnote{more info at: \url{https://www.sourceware.org/bzip2/}} algorithms with two different compression ratios, identified by -1 and -9. In particular, we can see that when using BZIP2, the pruned model is more than 20 times smaller than the unpruned model. The CPU inference time of the pruned model is $\sim0.001$s for one batch.

\begin{table*}[]
\centering
\caption{Test results for LeNet-5 architecture on Fashion-MNIST dataset. (var) is the variance computed over 10 runs.}
\label{tab:lent5-fMNIST-results}
\resizebox{\textwidth}{!}{%
\begin{tabular}{lllllccc}
\hline
                                                        & \multicolumn{4}{c}{Residual Weights (\%)} & $\text{Accuracy}_{(var)}$ (\%)  & Sparsity (\%)  & \multicolumn{1}{c}{\begin{tabular}[c]{@{}c@{}}Compression\\ Ratio\end{tabular}} \\
Methods                                                 & Conv1     & Conv2    & FC1     & FC2      &                &                &                                                                                \\
Baseline                                                & 100       & 100      & 100     & 100      & 91.90          & --             &                                                                                \\
                                                        &           &          &         &          &                &                &                                                                                \\
Tart. et al., 2018 \cite{tartaglione1} & 76.2      & 32.56    & 6.5     & 44.02    & 91.50          & 91.48          & 11.7x                                                                          \\
Han et al., 2015 \cite{Han2015}        & -         & -        & -       & -        & 91.56          & 93.04          & 14.3x                                                                          \\
Tart. et al., 2021 \cite{tartaglione2} & 78.6      & 26.13    & 2.88    & 32.66    & 91.53          & 95.70          & 23.3x                                                                          \\
Our method                                              & 78.84     & 17.84    & 1.20    & 6.26     & $\textbf{91.70}_{\:(0.08)}$ & \textbf{97.66} & \textbf{42.7x}                                                                 \\ \hline
\end{tabular}%
}
\end{table*}
    
\subsection{LeNet-5 on Fashion-MNIST}
\label{LeNet-5 on Fashion-MNIST}

We obtain the initial checkpoint after training for 21 epochs. We then use the hyperparameters shown in Table \ref{tab:image hyperparameters}, except for lower-bound = 90.5 and epochs = 75, for fine-tuning with regularization; finally, we fine-tune without regularization for 50 additional epochs. Table \ref{tab:lent5-fMNIST-results} compares performances on this dataset.

As can be seen, our method reaches the best performance both in terms of accuracy and compression. Our method is approximately 2 times better in compression rate than \cite{tartaglione2} with a 0.2\% accuracy improvement. Similarly, our results with LeNet-5 on MNIST are due to an effective sparsification of the fully connected layers. The disk space occupied by the pruned and unpruned model after compression using GZIP and BZIP2 is reported in Table \ref{tab:models_compression} and is comparable with what was obtained for the same architecture on MNIST. The CPU inference time of the pruned model is $\sim0.001$s for one batch.

\subsection{ResNet32 on CIFAR-10}
\label{ResNet32 on CIFAR-10}

This model \cite{He2016} is composed of a convolution layer with 16 3x3 filters, a batch normalization layer, 3 ResNet layers and a final dense layer, for a total of 464,154 trainable parameters. All the ResNet layers are composed of 5 ResNet blocks with different configurations: they all share 2 batch normalization layers but differ in the number of kernels generated by the 2 convolutional layers (16 3x3 filters for the blocks of the first ResNet layer, 32 3x3 for the second ResNet layer and 64 3x3 for the third ResNet layer). 

The initial checkpoint is obtained with the hyperparameters shown in  Table \ref{tab:image hyperparameters}, training for 200 epochs with $\eta = 0.1$.
After fine-tuning with regularization, we fine-tune without it for 1 last epoch with a batch size = 64 and $\eta$ = 6e-5. 

Table \ref{tab:resnet32-cifar10-results} shows that our technique for this more challenging task, which involves a deeper neural architecture and real images, outperforms all competitors in terms of sparsity. With respect to accuracy, both our method and \cite{tartaglione2} reach the baseline values, but our method improves sparsification by $1.45\%$ over \cite{tartaglione2}. In addition, the results show that in
the case of a complex deep network with residual layers such as ResNet-32, it is possible to prune a large percentage of weights without loss in classification performance and without any change in the pruning algorithm.  The disk space occupied by the pruned and unpruned model after compression using GZIP and BZIP2 is reported in Table \ref{tab:models_compression}; in particular we can see that using BZIP2 the pruned model is 4 times smaller on disk than the unpruned model. The CPU inference time of the pruned model is $\sim0.011$s for one batch.

\begin{table}[]
\centering
\caption{ResNet-32 on CIFAR-10. (var) is the variance computed over 10 runs.}
\label{tab:resnet32-cifar10-results}
\resizebox{\linewidth}{!}{%
\begin{tabular}{lccc}
\hline
Methods                                                 & $\text{Accuracy}_{(var)}$ (\%)  & Sparsity (\%)  & \begin{tabular}[c]{@{}c@{}}Compression\\ Ratio\end{tabular} \\
Baseline                                                & 92.67          & --             & --                                                         \\
                                                        &                &                &                                                            \\
Sparse VD \cite{Molchanov17}           & 92.12          & 50.11          & 2.0x                                                       \\
L0 \cite{Louizos18}                    & 91.20          & 60.00          & 2.5x                                                       \\
Han et al., 2015 \cite{Han2015}        & 91.92          & 71.51          & 3.51x                                                      \\
Targeted Dropout \cite{Gomez19}        & 92.54          & 80.00          & 5.0x                                                       \\
Tart. et al., 2021 \cite{tartaglione2} & \textbf{92.67} & 80.11          & 5.0x                                                       \\
Our method                                              & $\textbf{92.67}_{\:(0.01)}$ & \textbf{81.27} & \textbf{5.33x}                                             \\ \hline
\end{tabular}%
}
\end{table}

\subsection{ResNet50 on ImageNet}
\label{ResNet50 on ImageNet}

\begin{table*}[]
\centering
\caption{Test results for the ResNet50 architecture on ImageNet dataset. (var) is the variance computed over 10 runs. The upper part of the table shows the results for  compression ratios up to 2.5x, while the bottom part shows the results for ratios greater than 2.5x.}
\label{tab:resnet50-imagenet-results}
\resizebox{\textwidth}{!}{%
\begin{tabular}{lcccc}
\hline
Method         & \multicolumn{1}{l}{Baseline Accuracy (\%)} & \multicolumn{1}{r}{$\text{Accuracy}_{(var)}$ (\%)} & \multicolumn{1}{r}{Sparsity (\%)} & \begin{tabular}[c]{@{}c@{}}Compression\\ Ratio\end{tabular} \\ \hline
SSS-32, 2018 \cite{Huang18}   & 76.15                                      & 74.18                             & 27.01                             & 1.37x                                                      \\
OED, 2019 \cite{Wang19}     & 76.15                                      & 74.35                             & 23.67                             & 1.31x                                                      \\
Asympto, 2020 \cite{YHe20} & 76.15                                      & 75.53                             & --                                & --                                                         \\
Hrank, 2020 \cite{MLin20}   & 76.15                                      & 74.98                             & 36.71                             & 1.58x                                                      \\
DCP, 2018 \cite{Zhuang18}     & 76.01                                      & 74.95                             & 51.46                             & 2.06x                                                      \\
PKPSMIO, 2022 \cite{Zhu22a} & 76.09                                      & 75.86                             & 56.15                             & 2.28x                                                      \\
SSR-L2,1, 2002 \cite{Lin20} & 76.16                                      & 73.95                             & 37.79                             & 1.6x                                                       \\
SSR-L2,0, 2020 \cite{Lin20} & 76.16                                      & 74.00                             & 39.36                             & 1.64x                                                      \\
Our            & 76.16                                      & $\textbf{75.96}_{\:(0.001)}$                 & \textbf{59.13}                    & \textbf{2.44x}                                             \\ \hline
PKPSMIO, 2022 \cite{Zhu22a} & 76.09                                      & 74.61                             & 72.23                             & 3.60x                                                      \\
Our            & 76.16                                      & $\textbf{74.67}_{\:(0.003)}$                    & \textbf{73.85}                    & \textbf{3.82x}                                             \\ \hline
\end{tabular}%
}
\end{table*}

\begin{table*}[]
\centering
\caption{Model dimensions on disk.}
\label{tab:models_compression}
\resizebox{\textwidth}{!}{%
\begin{tabular}{ll|rr|rr}
\hline
\multicolumn{2}{c|}{\multirow{2}{*}{Model}} & \multicolumn{2}{c|}{Gzip}                                  & \multicolumn{2}{c}{bzip2}                                   \\
\multicolumn{2}{c|}{}                       & \multicolumn{1}{c}{Gzip -1} & \multicolumn{1}{c|}{Gzip -9} & \multicolumn{1}{c}{bzip2 -1} & \multicolumn{1}{c}{bzip2 -9} \\ \hline
LeNet5 on MNIST (1.7MB)       & Non-Pruned  & 1.6 MB                      & 1.6 MB                       & 1.7 MB                       & 1.6 MB                       \\
                              & Pruned      & 0.24 MB                     & 0.09 MB                      & 0.08 MB                      & 0.07 MB                      \\
LeNet5 on F-MNIST (1.7MB)     & Non-Pruned  & 1.6 MB                      & 1.6 MB                       & 1.7 MB                       & 1.6 MB                       \\
                              & Pruned      & 0.28 MB                     & 0.13 MB                      & 0.11 MB                      & 0.10 MB                      \\
ResNet32 on CIFAR-10 (1.9MB)  & Non-Pruned  & 1.8 MB                      & 1.7 MB                       & 1.8 MB                       & 1.8 MB                       \\
                              & Pruned      & 0.62 MB                     & 0.48 MB                      & 0.44 MB                      & 0.43 MB                      \\
ResNet50 on ImageNet  (98MB)  & Non-Pruned  & \multicolumn{1}{c}{91MB}    & \multicolumn{1}{c|}{91MB}    & \multicolumn{1}{c}{95MB}     & \multicolumn{1}{c}{93MB}     \\
                              & Pruned      & \multicolumn{1}{c}{41MB}    & \multicolumn{1}{c|}{33MB}    & \multicolumn{1}{c}{30MB}     & \multicolumn{1}{c}{30MB}     \\ \hline
\end{tabular}%
}
\end{table*}

The model comprises a convolutional layer with batch normalization and max pooling followed by 4 ResNet layers and a final average pooling layer with a fully connected classifier on top. All ResNet layers are composed of a different number of ResNet blocks, which comprise 3 convolutional layers interleaved with 3 batch normalization layers. Most of the convolutional layers mostly have 3×3 filters and downsampling is performed at the end of every first ResNet block relying on convolution with stride = 2. The networks contain 25,557,032 weights.
For our experiments, we start the regularized finetuning from a pretrained checkpoint\footnote{https://pytorch.org/vision/stable/index.html} using
the hyperparameters shown in Table \ref{tab:image hyperparameters}. We perform further fine-tuning for 5 epochs to obtain the best accuracy. A comparison between the results of our model and those of recent competitor works is shown in Table \ref{tab:resnet50-imagenet-results}. Our technique is shown to be very effective on ResNet50, achieving up to 3.83x sparsity with respect to the baseline and with only a slight loss in accuracy ($\sim 1.9\%$). Moreover, our solution is effective in both low and high compression regimes (i.e., $<2.5$x and $\ge2.5$x), resulting in the best accuracy and sparsity levels when compared to those of all other techniques. The amount of disk space occupied by the pruned and unpruned models after compression using GZIP and BZIP2 are reported in Table \ref{tab:models_compression}. Notably, when using BZIP2, the pruned model requires more than 3 times less disk space than the unpruned one. The CPU inference time of the pruned model is $\sim2.09$s for one batch.

\section{Language Generation}
\label{sec:language}

The first studies concerning sparsity in language architectures appeared recently \cite{Molchanov17,Gale2019}, following the progressive replacement of recurrent architectures by transformer-based models, and mainly focused on attention head pruning \cite{Michel2019,Voita2019}.
Nonetheless, with respect to the large number of available sparsity techniques for imaging, it is rare to find as many results in the language generation context.
We try to fill this gap by sparsifying the transformer \cite{Vaswani2017} on two different language tasks: dialog learning and machine translation. 

We implemented the model using the HuggingFace\footnote{https://huggingface.co/} library which provides easy access to different datasets, tokenizers and output generation techniques.
Since our sparsification algorithm is not architecture specific, no modifications are needed with respect to what is described in Sec. \ref{general_approach_pruning}.

In addition, as previously mentioned in the same section,  we use the BLEU (Bilingual Evaluation Understudy) score, which compares a generated sentence to a reference sentence, as a suitable metric for evaluation in the language generation context. It works by counting matching n-grams in the candidate translation to n-grams in the reference text. A perfect match results in a score of 1.0, whereas a perfect mismatch results in a score of 0.0. BLEU was originally proposed by  \cite{Papineni2002} for evaluating the predictions made by automatic machine translation systems but is now commonly used for many other language generation tasks such as dialog generation, image caption generation, text summarization and speech recognition.
\subsection{Dataset Description}

\paragraph{WMT14} WMT14 \cite{bojar2014} is a collection of datasets presented in the Ninth Workshop on Statistical Machine Translation. It comes as a parallel corpus made by sentences translated into various languages. 
It is derived from many different sources, among which there are the Europarl corpus \cite{koehn2005epc} (created from the European Parliament Proceedings in the official languages of the EU), the News Commentary \cite{TIEDEMANN12.463} corpus and the Common Crawl corpus \cite{CommonCrawl} (which was collected from web sources).

 For our experiments we use the English to German translation dataset En-De WMT14, provided by the Stanford Natural Language Processing Group \cite{SNLP}, which is more than 300x larger than Taskmaster-1; it contains 4,468,840 training samples and 3000 test samples. Some source-translation examples are shown in Table \ref{tab:translation-pairs-En-De}.

\paragraph{Taskmaster-1} 
Taskmaster-1 \cite{byrne2019} is a public crowdsourced dataset, released by Google in 2019, where Amazon Turkers were asked to write dyadic dialogs (see Table \ref{tab:Dialog sample from Taskmaster-1}) following some given set of instructions describing six tasks: ordering pizza, creating autorepair appointments, setting up rides for hire, ordering movie tickets, ordering coffee drinks and making restaurant reservations. 
The dataset is composed of 13,215 task-based dialogs (12,355 for the training set and 770 for the test set), including 5,507 spoken and 7,708 written dialogs. 
    
\begin{table*}
\centering
\caption{Example of translation pairs from En-De WMT14.}
\label{tab:translation-pairs-En-De}
\resizebox{\textwidth}{!}{%
\begin{tabular}{l}
\hline
\multicolumn{1}{c}{\textbf{\begin{tabular}[c]{@{}c@{}}Source\\ Translation\end{tabular}}}                                                                                                                             \\ \hline
\begin{tabular}[c]{@{}l@{}}Iron cement protects the ingot against the hot, abrasive steel casting process.\\ Nach der Aushärtung schützt iron cement die Kokille gegen den heissen, abrasiven Stahlguss.\end{tabular} \\ \hline
\begin{tabular}[c]{@{}l@{}}Goods and services advancement through the P.O.Box system is NOT ALLOWED. \\ der Vertrieb Ihrer Waren und Dienstleistungen durch das Postfach System WIRD NICHT ZUGELASSEN.\end{tabular}   \\ \hline
\begin{tabular}[c]{@{}l@{}}Their achievement remains one of the greatest in recent history. \\ Das bleibt eine der größten Errungenschaften in der jüngeren Geschichte.\end{tabular}                                  \\ \hline
\end{tabular}%
}
\end{table*}

\begin{table*}
\caption{Dialog sample from Taskmaster-1.}
\label{tab:Dialog sample from Taskmaster-1}
\resizebox{\textwidth}{!}{%
\begin{tabular}{l}
\toprule
\textbf{Input} \tabularnewline
\midrule
\textless user\textgreater Hi there, could you please help me with an order of Pizza?\textless enduser\textgreater \\ \textless agent\textgreater Sure, where would you like to order you pizza from?\textless endagent\textgreater \\ \textless user\textgreater I would like to order a pizza from Domino's.\textless enduser\textgreater \\ \textless agent\textgreater What kind of pizza do you want to order? \textless endagent\textgreater \\ \textless user\textgreater What are the specials they have right now? \textless enduser\textgreater \\ \textless agent\textgreater There are family and party combos currently on offer \textless endagent \textgreater \\ \textless user\textgreater No, I just want a large pizza \textless enduser\textgreater \\ \textless agent\textgreater They have any large specialty pizza for 10.99 \textless endagent\textgreater \\ \textless user\textgreater What are their specialty pizzas? \textless enduser\textgreater
\tabularnewline
\midrule
\textbf{Target}
\tabularnewline
\midrule
\textless agent\textgreater Well, there is the Extravagazza, Meatzza, Philly Cheesesteak,\\ Hawaiian, Buffalo Chicken Ranch, and more. Would you like to hear more? \textless endagent\textgreater
\\ \bottomrule
\end{tabular}%
}
\end{table*}

\begin{table}
\caption{Transformer hyperparameters.}
\label{tab:language hyperparameters}
\begin{tabular}{lll}
Hyperparameters    & Taskmaster-1 & WMT14 \\ \hline
vocabulary      & 32k & 32k \\
\# encoder layers     & 2 & 6 \\
\# decoder layers     & 2 & 6 \\
\# attention heads    & 4 & 8 \\
feed forward dim    & 256 & 2048 \\
embedding dim.    & 256 & 512 \\
\# weights   & 10M & 61M  \\
max sequence len. & 256 & 256 \\
beam size & 6 & 4 \\
length penalty & - & 0.6 \\
$\eta$ & 0.0005 & 2.5e-5\\
Optimizer & Adam & Adam\\
batch size & 150 & 120\\
lower-bound & 6.03 & 27.3\\
$\lambda$ & 1e-5 & 2.22e-7\\
pruning-percentage & 0.1 & 0.1\\
eval-interval  & 300 & 6000\\
\end{tabular}
\end{table}

\subsection{Transformer on WMT14}
\label{WMT14_Exp}
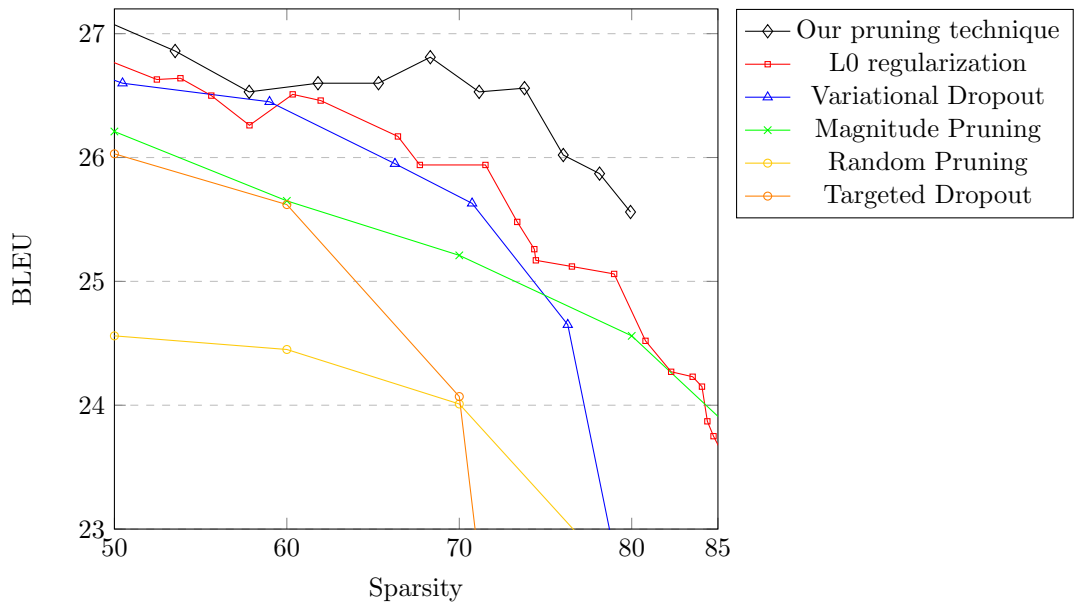
\begin{figure*} [h]
\centering
    \begin{tikzpicture}[]
        \begin{axis}[
            title={},
            xlabel={Sparsity},
            ylabel={BLEU},
            xmin=50, xmax=85,
            ymin=23, ymax=27.2,
            xtick={50,60,70,80,85},
            ytick={23,24,25,26,27,28},
            legend pos=outer north east,
            ymajorgrids=true,
            grid style=dashed,
            scale only axis,
            width=0.5\textwidth,
        ]
        
        \addplot[
            color=black,
            mark=diamond,
            mark size=2.5pt
            ]
            coordinates {
                (48.71,27.15)(53.52100000000001,26.86)(57.8137,26.53)(61.806,26.6)(65.315,26.6)(68.32,26.81)(71.1511,26.53)(73.7825,26.56)(76.0307,26.02)(78.1347,25.87)(79.9308,25.56)
            };
        
        \addplot[
            color=red,
            mark=square,
            mark size=1pt 
            ]
            coordinates {
            (49.5496064425,26.79)(52.461797,26.63)(53.82964610000001,26.64)(55.6197405,26.5)(57.8327179,26.26)(60.3384554,26.51)(61.96107270000001,26.46)(66.442591,26.17)(67.7134633,25.94)(71.51768209999999,25.94)(73.3617902,25.48)(74.3513167,25.26)(74.4411349,25.17)(76.53173209999999,25.12)(78.980428,25.06)(80.7998478,24.52)(82.2867811,24.27)(83.5332572,24.23)(84.0732813,24.15)(84.3871653,23.87)(84.7429812,23.75)(86.1137569,23.38)(86.76145670000001,23.54)(86.9241893,23.14)(88.2563233,22.7)(88.9639258,22.55)(89.0202522,22.97)(90.5870318,22.14)(90.982759,21.23)(91.9579148,21.48)(92.55552290000001,21.1)(93.3664203,21.24)(94.1744685,20.6)(94.5438683,19.85)(95.4818785,19.69)(95.95679640000001,19.39)(96.7341006,18.81)
            };
            
        \addplot[
            color=blue,
            mark=triangle,
            mark size=2pt
            ]
            coordinates {
                (48.4709551334,26.7)(50.4709551334,26.6)(58.9960038662, 26.45)(66.270262,25.95)(70.74294090000001,25.63)(76.2913585,24.65)(79.55823540000001,22.42)(79.57976459999999,19.42)(79.72303629999999,20.04)(79.7921419,15.94)(80.31461240000001,20.81)
            };

         \addplot[
            color=green,
            mark=x,
            mark size=2pt
            ]
            coordinates {
            (50,26.21)(59.9990845,25.65)(70.0000525,25.21)(80.0001085,24.56)(90.00006909999999,23.26)(95.00006440000001,20.75)(98.00012709999999,16.37)(98.00012709999999,16.05)
            };
        
        \addplot[
            color=yellow!80!red,
            mark=o,
            mark size=1.5pt
            ]
            coordinates {
                (50,24.56)(60.00014539999999,24.45)(70.0000405,24.01)(79.99945279999999,22.48)(80.0001025,22.92)(90.00006909999999,20.67)(95.0000525,17.23)(98.00012709999999,10.22)

            };
        \addplot[
            color=orange!80!orange,
            mark=o,
            mark size=1.5pt
            ]
            coordinates {
                (50,26.03)(60.000,25.62)(70.000,24.07)(80.000,12.39)(90.0000,0.07)

            };

        \legend{Our pruning technique, L0 regularization, Variational Dropout, Magnitude Pruning, Random Pruning, Targeted Dropout}
        \end{axis}
    \end{tikzpicture}
    \caption{BLEU results comparison at different sparsity levels on the WMT14 dataset. Except for our technique, the datapoints are taken from \cite{Gomez19} for targeted dropout and from \cite{Gale2019} for the others methods, for which we take only their best runs (in terms of BLEU).\newline}
    \label{fig:sparsity_wmt}
    \end{figure*}

\begin{table*}[]
\centering
\caption{Test results for Transformer architecture on WMT14 dataset. (var) is the variance computed over 5 runs.}
\label{tab:WMT14results}
\resizebox{\textwidth}{!}{%
\begin{tabular}{l|lllll|ccc}
\hline
          & \multicolumn{5}{c|}{Residual Weights (\%)}                      &       &               &                                                            \\
Model     & Encoder   & Decoder   & Encoder & Decoder & Embedding/          & $\text{BLEU}_{(var)}$  & Sparsity (\%) & \begin{tabular}[c]{@{}c@{}}Compression\\ Ratio\end{tabular} \\
          & Attention & Attention & FFN     & FFN     & Classification Head &       &               &                                                            \\
Baseline  & 100       & 100       & 100     & 100     & 100                 & 27.20 & --            & --                                                         \\
Our model & 24.70     & 27.54     & 21.27   & 20.10   & 12.43               & $25.56_{\:(0.01)}$ & \textbf{79.93}         & \textbf{4.8x}                                                       \\ \hline
\end{tabular}%
}
\end{table*}

Details regarding the transformer architecture are given in Table \ref{tab:language hyperparameters} and follow the settings from \cite{Gale2019}.

To obtain the initial checkpoint, we train the model for 10 epochs with batch size = 120 and  $\eta=5e-05$, using the Adam optimizer $(\beta_1 =0.85, \beta_2=0.997, eps = 1e-8)$ and achieve BLEU performance comparable to the baseline defined in \cite{Gale2019}.

Starting from the checkpoint described above, the process of fine-tuning with regularization continues for 16 epochs with batch size = 100, $\eta=2.5e-05$ and  $\lambda =  2.22e-07$. Evaluations on the validation set are carried out every 6000 steps (24 times each epoch) with BLEU lower-bound = 27.3 and 10\% of the remaining weights pruned when required.
Finally we finetune without regularization for 5 more epochs. With respect to \cite{Gale2019}, we stop pruning when the validation performance reaches a plateau (or suddenly declines) and never surpasses the user-defined lower-bound, as described in Section \ref{general_approach_pruning}. This criterion causes the pruning to stop at $\sim$80 \% sparsity.

As shown in Figure~\ref{fig:sparsity_wmt}, our pruning technique performs better than all other methods, preserving BLEU values up to $\sim$75\% sparsity while dropping at most 0.5 points with respect to the baseline.

It also seems to be more resilient at higher compression levels since the BLEU scores start to degrade visibly only after $\sim$75 \% sparsity is reached, whereas those of the other five pruning methods degrade earlier. 
This is the first experiment where our technique is used in the context of natural language generation, showing that it is very generalizable and can be effectively applied to transformer models that are heavily based on attention layers and present many shared weights, such as in the word embedding layer and in the attention layer itself.

Table~\ref{tab:WMT14results} shows in detail the layer-by-layer and global pruning percentages at the higher compression level reached. 
Table \ref{tab:model-size-wmt14} shows the amount of disk space occupied by the pruned and unpruned models after being compressed using GZIP and BZIP2. Notably, when using BZIP2, the pruned model requires 4 times less disk space than the unpruned one. The CPU inference time of the pruned model is $\sim4.05$s for one batch.

    \begin{table}
        \caption{Disk space dimensions of pruned/unpruned transformer models on WMT14.}
        \label{tab:model-size-wmt14}
        \centering
        \resizebox{0.48\textwidth}{!}{%
        \begin{tabular}{l|rr|rr}
        \hline\noalign{\smallskip}
        \textbf{Model} & \multicolumn{2}{c|}{\textbf{GZIP}} & \multicolumn{2}{c}{\textbf{BZIP2}} \\
        246.5 MB & \multicolumn{1}{c}{GZIP -1} & \multicolumn{1}{c|}{GZIP -9} & \multicolumn{1}{c}{BZIP2 -1} & \multicolumn{1}{c}{BZIP2 -9} \\
        \noalign{\smallskip}\hline
        \noalign{\smallskip}
        Unpruned & 228.9 MB & 228.3 MB & 237.1 MB & 233.2 MB \\
        Pruned & 87.2 MB & 66.8 MB & 59.9 MB & 58.2 MB \\
        \noalign{\smallskip}\hline
        \end{tabular}%
        }
    \end{table}

\subsection{Transformer on Taskmaster-1}
\label{Tsk1_Exp}

\begin{figure*} [h]
\centering
    \begin{tikzpicture}[]
        \begin{axis}[
            title={},
            xlabel={Sparsity},
            ylabel={BLEU},
            xmin=48, xmax=95,
            ymin=5.9, ymax=6.8,
            xtick={50,60,70,80,90},
            ytick={5.9,6.0,6.1,6.2,6.3,6.4,6.5,6.6,6.7},
            legend pos=outer north east,
            ymajorgrids=true,
            grid style=dashed,
            scale only axis,
            width=0.5\textwidth,
        ]
        
        \addplot[
            color=black,
            mark=diamond,
            mark size=2.5pt
            ]
            coordinates {
                (49.1059,6.58)(53.07280000000001,6.61)(57.0417,6.43)(61.0948,6.21)(64.56770000000001,6.54)(67.86760000000001,6.260000000000001)(70.9012,6.39)(73.6529,6.35)(76.1539,6.510000000000001)(78.2842,6.140000000000001)(80.3159,6.18)(82.1615,6.02)(83.8131,6.39)(85.3434,6.16)(86.6984,6.2700000000000005)(87.9429,6.140000000000001)(89.0509,6.0600000000000005)(91.29,6.260000000000001)
            };
        
        \addplot[
            color=red,
            mark=square,
            mark size=1pt 
            ]
            coordinates {
            (46.50909999999999,6.11)(56.6488,6.21)(65.2163,6.140000000000001)(72.2553,6.260000000000001)(78.09809999999999,6.0600000000000005)(82.59369999999999,6.13)(85.9936,6.54)(88.6477,6.18)(90.74,6.22)
            };
            
        \addplot[
            color=blue,
            mark=triangle,
            mark size=2pt
            ]
            coordinates {
                (45.7276,6.39)(55.49090000000001,6.15)(63.7636,6.550000000000001)(70.6331,6.069999999999999)(76.2515,6.2)(80.7119,5.99)(84.3317,5.99)(87.24959999999999,6.1)(90.74,6.1899999999999995)
            };

        \legend{Our pruning technique, L1 regularization, L2 regularization}
        \end{axis}
    \end{tikzpicture}
    \caption{BLEU results comparison at different sparsity levels on Taskmaster-1 dataset. \newline}
    \label{fig:sparsity_tskm}
    \end{figure*}
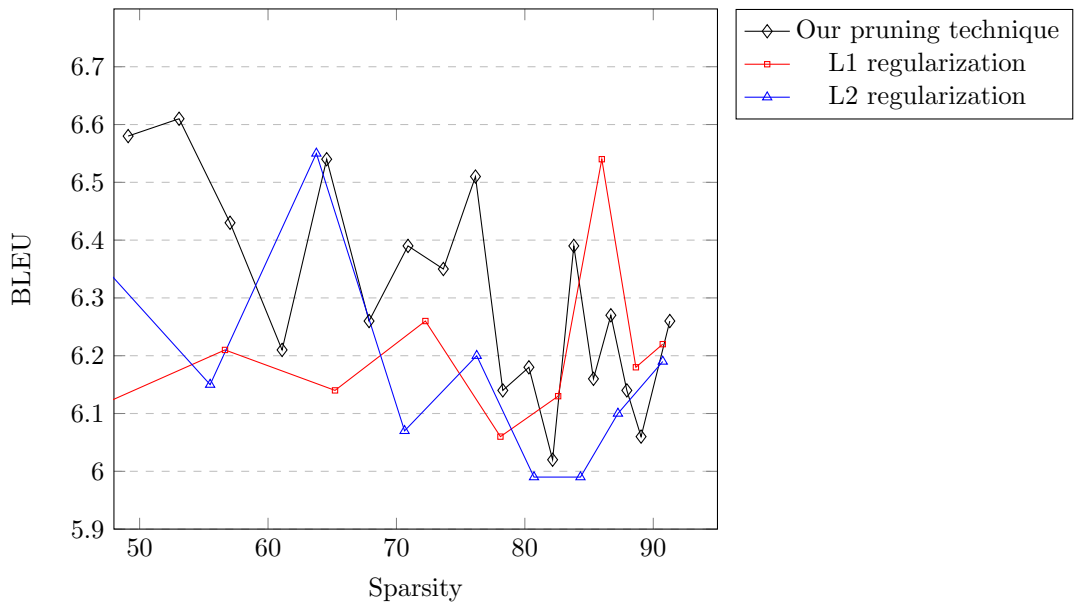

\begin{table*}[]
\centering
\caption{Test results for Transformer architecture on Taskmaster-1 dataset. (var) is the variance computed over 10 runs.}
\label{tab:tsk-results}
\resizebox{\textwidth}{!}{%
\begin{tabular}{l|lllll|ccc}
\hline
          & \multicolumn{1}{c}{Residual Weights (\%)} &           &         &         &                     &               &                & \multicolumn{1}{l}{}                                       \\
Model     & Encoder                                   & Decoder   & Encoder & Decoder & Embedding/          & $\text{BLEU}_{(var)}$          & Sparsity (\%)  & \begin{tabular}[c]{@{}c@{}}Compression\\ Ratio\end{tabular} \\
          & Attention                                 & Attention & FFN     & FFN     & Classification Head &               &                & \multicolumn{1}{l}{}                                       \\ \hline
Baseline  & 100                                       & 100       & 100     & 100     & 100                 & 6.11          & --             & --                                                         \\
L1        & 2.53                                      & 18.57     & 1.17    & 53.08   & 20.99               & $6.22_{\:(0.02)}$          & 90.74          & 10.8x                                                       \\
L2        & 17.71                                     & 21.20     & 15.21   & 26.61   & 6.28                & $6.18_{\:(0.02)}$         & 90.74          & 10.8x                                              \\
Our model & 21.53                                     & 21.77     & 21.68   & 26.93   & 7.12                & $\textbf{6.26}_{\:(0.02)}$ & \textbf{91.29}          & 11.5x                                                      \\ \hline
\end{tabular}%
}
\end{table*}
    
Following \cite{byrne2019}, we use the dialog context up to the last user turn as input-data, and as a target for the subsequent assistant utterance. An example of this format is shown in Table \ref{tab:Dialog sample from Taskmaster-1}.

Transformer architecture details are presented in \cite{byrne2019}, and shown in Table \ref{tab:language hyperparameters}. We train it for 15 epochs using the Adam optimizer $(\beta_1 =0.85, \beta_2=0.997, eps = 1e-8)$ with batch size $= 150$ and dropout = 0.2.
The final checkpoint we obtain shows comparable BLEU performance to the author's model.

Starting from this checkpoint, we fine-tune with regularization for 40 epochs with $\eta=0.0005$. We rely on a small $\lambda = 1e-05$ to avoid losing performance during the early stages. We find that checking every 300 training steps (i.e., 4 times every epoch) is a good compromise to obtain frequent pruning steps while retaining generation ability.
The BLEU lower-bound is set to 6.03, which is very close to the author's baseline result of 6.11, and the pruning-percentage is 10\%.
After 30 epochs, the algorithm makes the last pruning, and the last 10 epochs are used to recover the BLEU score.

To the best of our knowledge, there are no achievements yet in the literature about weight sparsity in dialog generation tasks. We, therefore, establish the first results in this context, displayed in Table \ref{tab:tsk-results}, testing our method against L1 and L2 regularization baselines.   

Our sparsification technique allows us to obtain a highly sparsified model, with a sparsity level greater than 90\%. Moreover, our final BLEU is even higher than the original result, suggesting that in some cases a sparsified model can generalize better than a nonsparsified model. In Figure \ref{fig:sparsity_tskm}, we show the BLEU scores of our technique and the L1 and L2 baselines at different sparsity levels. In this case, the gap between our method and the others is less evident due to the high variability of the BLEU scores. This is probably given by the fact that Taskmaster-1 contains rather short sentences when compared to the WMT-14 dataset, so even small output differences with the target sentence have a high impact on the final score, which is based on n-gram counting. Regardless, our system is almost always able to perform better than the L1 and L2 regularizations with the exception of sparsity levels between 0.8 and 0.9, where L1 is preferred.

\begin{table}
        \caption{Disk space dimensions of pruned/unpruned transformer models on Taskmaster-1.}
        \label{tab:model-size-TSK}
        \centering
        \resizebox{0.48\textwidth}{!}{%
        \begin{tabular}{l|rr|rr}
        \hline\noalign{\smallskip}
        \textbf{Model} & \multicolumn{2}{c|}{\textbf{GZIP}} & \multicolumn{2}{c}{\textbf{BZIP2}} \\
        43.5 MB & \multicolumn{1}{c}{GZIP -1} & \multicolumn{1}{c|}{GZIP -9} & \multicolumn{1}{c}{BZIP2 -1} & \multicolumn{1}{c}{BZIP2 -9}\\
        \noalign{\smallskip}\hline
        \noalign{\smallskip}
        Unpruned & 40.2 MB & 40.1 MB & 41.7 MB & 41.1 MB \\
        Pruned & 11.2 MB & 7.4 MB & 6.4 MB & 6.2 MB \\
        \noalign{\smallskip}\hline
        \end{tabular}%
        }
    \end{table}
    
Table \ref{tab:model-size-TSK} shows the disk space occupied by the pruned and unpruned models after being compressed. Additionally, in this case very good compressions can be achieved. In particular, using BZIP2, the pruned model is approximately 6,6 times smaller on the disk than the unpruned model. The CPU inference time of the pruned model is $\sim2.28$s for one batch.

\section{Conclusions}
The identification of irrelevant model parameters for pruning is the focal point of this work. We propose a solution that is an improvement of classical weight decay and consequently suitable for any functional loss. Moreover, it is simple to implement and results in a largely usable and general framework that proves to be effective in sparsifying different deep architectures.

We reach state-of-the-art results in one out of four image classification datasets and improve state-of-the-art for the others in terms of the combination of sparsity and accuracy, also obtaining a new state-of-the-art in the machine translation dataset WMT14. Since there are very few results for sparsity in language generation tasks, another contribution of this paper is that we give a new data point on Taskmaster-1.

A future interesting contribution can be to explore applications of our method to low-resource devices such as smartphones and IoT systems.

\bmhead{Acknowledgments}

The activity has been partially carried on in the context of the Visiting
Professor Program of the Gruppo Nazionale per il Calcolo Scientifico
(GNCS) of the Italian Istituto Nazionale di Alta Matematica
(INdAM).




%

\end{document}